\pdfoutput=1
\documentclass[11pt,letterpaper]{article}
\usepackage[margin=1in]{geometry}
\usepackage[T1]{fontenc}
\usepackage[utf8]{inputenc}
\usepackage{newtxtext}
\usepackage{amsmath}
\usepackage{newtxmath}
\usepackage{cite}
\usepackage{graphicx}
\usepackage{booktabs}
\usepackage{siunitx}
\usepackage{xcolor}
\usepackage{textcomp}
\usepackage{float}
\usepackage{microtype}
\usepackage{authblk}
\usepackage[colorlinks=true,urlcolor=blue,linkcolor=black,citecolor=black]{hyperref}
\hypersetup{pdftitle={Elastra: A Simulation Framework for Elastic Bodies},pdfauthor={Youran Sun, Chugang Yi, Xingyu Ren, Haizhao Yang}}

\graphicspath{{figs/}}
\makeatletter
\def\bstctlcite{\@ifnextchar[{\@bstctlcite}{\@bstctlcite[@auxout]}}
\def\@bstctlcite[#1]#2{\@bsphack
  \@for\@citeb:=#2\do{\edef\@citeb{\expandafter\@firstofone\@citeb}\if@filesw\immediate\write\csname #1\endcsname{\string\citation{\@citeb}}\fi}\@esphack}
\makeatother

\title{\LARGE\bfseries Elastra: A Simulation Framework for Elastic Bodies}
\author[1,*]{Youran Sun}
\author[1,*]{Chugang Yi}
\author[2]{Xingyu Ren}
\author[1]{Haizhao Yang}
\affil[1]{University of Maryland, College Park\quad \{sun1245, chugang, hzyang\}@umd.edu}
\affil[2]{The Chinese University of Hong Kong\quad xy.ren@link.cuhk.edu.hk}
\affil[*]{Equal contribution.}
\date{}

\begin{document}
\bstctlcite{IEEEexample:BSTcontrol}
\maketitle

\begin{figure}[!ht]
\centering
\includegraphics[width=\textwidth]{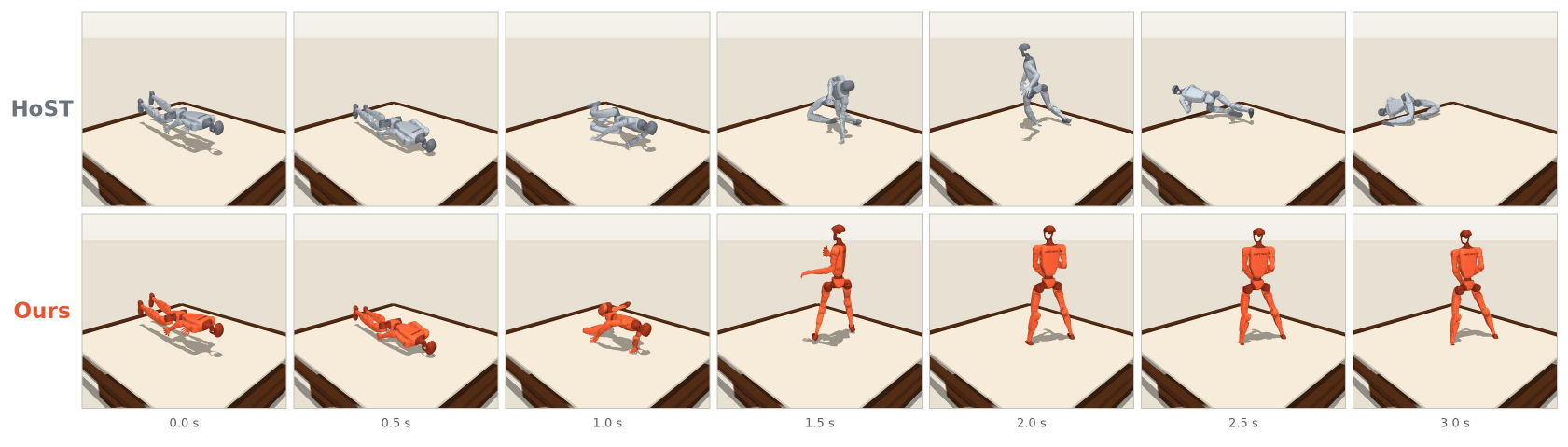}
\caption{The Unitree G1 stands up from a prone pose on a mattress simulated with Elastra, with a residual controller adapted from HoST.
The mattress sinks under each contact and recovers as the load moves.
Trampoline get-up is reported in Fig.~\ref{fig:host}.}
\label{fig:teaser}
\end{figure}

\begin{abstract}
When robots interact with elastic surfaces such as mattresses and trampolines, changes in loading cause the surfaces to deform and rebound, altering subsequent contact locations and forces.
We present Elastra, a simulation framework for modeling this interaction and supporting the training and evaluation of robot control policies.
Elastra represents each surface using elements that move only vertically, with elastic connections and damping to model deformation and recovery.
Contact forces drive surface motion, and the updated surface geometry is used to compute subsequent contact forces.
We systematically validate our simulations against an independent implementation and a real mattress load curve.
We further use Elastra to train the Unitree G1 humanoid to stand up on mattresses and trampolines.
Using the pretrained HoST policy and ProtoMotions motion tracker as original controllers, we train residual policies through reinforcement learning to add corrections to their joint targets.
Adding these residual policies increases the success rate with HoST from 44.1\% to 72.5\%, and from 55\% to 89\% on the trampoline.
With ProtoMotions, the success rate increases from 19.0\% to 45.0\% for initial poses not encountered during training.
The code and video are available on \href{https://github.com/yichg/elastra}{GitHub}.
\end{abstract}

\section{Introduction}
\label{sec:intro}

To use the motion skills of a robot in different environments, one has to consider how changed contact conditions affect the execution of a motion.
Mattresses, sofa cushions and trampolines sink where they are pressed and recover when the load moves on, so every push changes where the next contact lands and what force it meets.
Robot learning environments usually treat the ground as rigid.
This paper builds Elastra, a simulation framework in which mattresses and trampolines deform under the robot and the deformed surface enters the contact computation.
As the demonstration task we use standing up after a fall (Fig.~\ref{fig:teaser}), which loads and unloads the surface with several body parts in turn.
HoST~\cite{huang2025learninghumanoidstandingupcontrol} and HumanUP~\cite{he2025learninggettinguppoliciesrealworld} have demonstrated humanoid standing up from many fallen postures and on varied terrain, and Singh et al.~\cite{singh2025robusthumanoidwalkingcompliant} achieved humanoid walking on a mattress and soft foam.
This paper instead simulates the support itself as a body that moves under the robot.

The contributions are \textbf{(1)} Elastra, a mattress and trampoline simulation framework for robot learning that incorporates local deformation, neighboring-region motion, and post-unloading recovery into the contact computation, \textbf{(2)} a systematic validation of Elastra against (i) an independent implementation and (ii) a real mattress load curve, and \textbf{(3)} a demonstration in which residual reinforcement learning on top of a pretrained standing policy enables the Unitree G1 to stand up on both mattresses and trampolines, evaluated across different stiffness levels and on fallen postures unseen during training.

\section{Related Work}
\label{sec:related}

\textbf{Simulation of deformable supports.} The outcome of a contact simulation depends on the shapes, materials and contact parameters of the bodies and on the numerical method that solves the equations of motion.
In robotics simulators, the PhysX deformable bodies of Isaac Sim~\cite{isaacsim51_deformables}, Warp~\cite{macklin2022warp}, the flex bodies of MuJoCo~\cite{todorov2012mujoco, mujoco_docs} and SOFA~\cite{faure2012sofa} represent soft volumes and cloth, and benchmarks such as SoftGym~\cite{lin2021softgymbenchmarkingdeepreinforcement} and PlasticineLab~\cite{huang2021plasticinelabsoftbodymanipulationbenchmark} use them for manipulation.
Elastra is built inside MuJoCo and uses its flex bodies only for collision geometry, not for their elasticity.
A flex body is a volumetric finite element model with linear elements, so the elements must be small enough to follow the deformation under each contact.
In the task of standing up on a mattress, the robot presses on the mattress with feet, hands and knees, and each contact loads a patch about 0.1 m across.
The compression under such a patch fades within about one patch width to the side and in depth, so the elements must be smaller than the patch in length, width and depth.
On the mattress we consider, elements of this size amount to $10^4$ degrees of freedom.
Elastra avoids this cost.
Each cell of Elastra keeps one vertical degree of freedom, and a bending term carries the lateral spreading of a load, such that the state of the surface is the vector of cell heights and 380 degrees of freedom describe the same mattress.
Every restoring and damping force follows an explicit physical law given in Section~\ref{sec:backend}, so the material law stays open to calibration by a real mattress.
Together, the low cost and the observable surface state make Elastra friendly to reinforcement learning, and the response calibrated against a real mattress prepares the transfer to hardware.
Sun et al.~\cite{sun2026effectiveparametersrealbehavior} describe such calibrated values as effective parameters that absorb the details omitted by a simulator and reproduce real behavior.

\textbf{Robots on soft ground.} Foam, soft pads, and similar ground deform under load, and a robot moving on them has to adapt to changes in foot placement and contact force.
Singh et al.~\cite{singh2025robusthumanoidwalkingcompliant} varied the shape and the softness of the ground in training, achieved walking on a mattress and soft cushion foam with a real HRP-5P, and studied aperiodic walking with a variable step frequency.
This paper computes, with movable cells and their connections, the surface motion caused by any body part that loads and unloads the support; standing-up exercises with hands, knees, torso, and feet in turn.

\textbf{Standing up.} A basic choice in the control of standing up is how to obtain the sequence of motions from lying to standing.
Kanehiro et al.~\cite{kanehiro2007gettingup} searched a graph of postures to connect the current posture to an existing recovery motion.
In physics based character animation, DeepMimic~\cite{peng2018deepmimicexampleguideddeepreinforcement} combines the imitation of a reference motion with a task reward and learns a feedback policy that executes the motion from the current state.
Motion trackers trained in frameworks such as ProtoMotions~\cite{tessler2024protomotions} bring this approach to robots.
Learned standing control has also progressed on real humanoid robots.
For full size humanoids, HoST~\cite{huang2025learninghumanoidstandingupcontrol} does not rely on a predefined standing trajectory.
HumanUP~\cite{he2025learninggettinguppoliciesrealworld} separates the discovery of the motion from the training of a deployable controller.
These studies treat the ground as given, and our paper builds the simulation of supports that deform and recover under the robot, and uses standing up as a demonstration.

\textbf{Residual reinforcement learning.} Residual reinforcement learning~\cite{johannink2019residual,silver2019residualpolicylearning} adds a learned correction to the output of an existing controller, so that the task is completed with the original motion ability while the motion is adjusted from the new interaction.
On legged and humanoid robots, reinforcement learning has likewise been combined with an existing controller~\cite{chen2023learningagilelocomotionadaptive,kamohara2025rlaugmentedadaptivemodelpredictive,jeon2025residualmpcblendingreinforcement,li2026acceleratingscalingmpcguidedreinforcement}.

\section{Mattress and Trampoline Models}
\label{sec:backend}

\begin{figure}[t]
\centering
\includegraphics[width=0.95\textwidth]{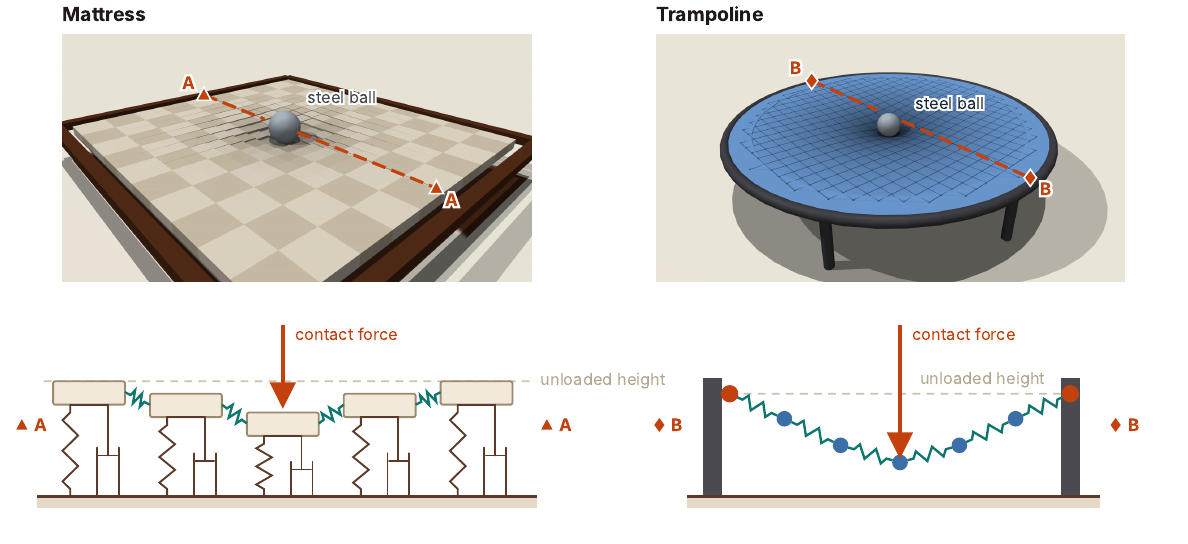}
\caption{The two surface models of Elastra.
Top: a steel ball resting on the simulated mattress (left) and trampoline (right); the dashed lines mark the sections A--A and B--B drawn below.
Bottom: mattress cells rest on local springs and dampers and are coupled to their neighbors through a bending term; trampoline nodes carry the load through the membrane connections to a fixed rim.
The renders use softer parameters than the experiments so that the deformation is visible.}
\label{fig:concept}
\end{figure}

We discretize the mattress and the trampoline in MuJoCo~\cite{todorov2012mujoco} into surface cells with mass, keep only the vertical translation of each cell, and describe deformation and recovery under load by elastic connections and damping.
Let the surface have $n_s$ movable nodes and let $x_i = z_i^{\mathrm{ref}} - z_i$ be the downward displacement of node $i$, where $z_i^{\mathrm{ref}}$ is its reference height and $z_i$ its current height.
Collecting the displacements of the movable nodes in a vector $x\in\mathbb R^{n_s}$, the nodes move according to
\begin{equation}
M_s\ddot x + C_s(\alpha)\dot x + f_{\mathrm{el}}(x;\alpha) = f_{\mathrm{contact}} + f_g .
\label{eq:eom}
\end{equation}
Here $M_s=\operatorname{diag}(m_1,\ldots,m_{n_s})$ is the node mass matrix, $C_s(\alpha)\dot x$ the damping force and $f_{\mathrm{el}}(x;\alpha)$ the elastic restoring force, and $\alpha>0$ is a dimensionless stiffness scale factor.
$f_{\mathrm{contact}}$ is the vertical component of the contact forces of the robot on the nodes and $f_g$ the weight of the nodes.

\textbf{Mattress.} The mattress is a rectangular grid of blocks above a rigid floor (Fig.~\ref{fig:concept}, left).
Each block resists local compression through a vertical spring and damper, and for every interior cell a bending term resists the curvature of the surface at that cell.
The mattress moves according to
\begin{equation}
m\ddot x + c\dot x + \left(k\,\mathbf I + b\,\mathbf L^{\top}\mathbf L\right)x = f_{\mathrm{contact}} + m g\,\mathbf 1 ,
\label{eq:cell}
\end{equation}
where $\mathbf L$ is the five point Laplacian over the interior cells, $(\mathbf L x)_i = x_N + x_S + x_E + x_W - 4x_i$, $k$ the cell stiffness, $b$ the bending stiffness, $c$ the cell damping and $m$ the cell mass, so that the elastic energy is $k|x|^2/2 + b|\mathbf L x|^2/2$.
The bending term spreads a local load over the neighboring cells; because every row of $\mathbf L$ sums to zero, the weight of the cells sags the unloaded bed uniformly by $mg/k$ and the surface the robot lands on is flat.
There is no separate connection between pairs of cells and no nonlinear compression term.
We change the stiffness of the mattress through the single factor $\alpha$,
\begin{equation}
k(\alpha)=\alpha k_1,\quad b(\alpha)=\alpha b_1,\quad c(\alpha)=\sqrt{\alpha}\,c_1,
\label{eq:scaling}
\end{equation}
where $k_1$, $b_1$ and $c_1$ are the reference values, so that the damping ratio of a cell and the shape of the dent under a load are the same at every stiffness and only its depth changes.
Geometry, mass and contact parameters stay fixed when the stiffness changes.

\textbf{Trampoline.} The trampoline is a circular network of nodes with a fixed rim, and its interior nodes translate vertically (Fig.~\ref{fig:concept}, right).
Neighboring nodes transfer load through connections within the membrane, and no interior node has its own spring to the ground.
For an interior node $i$ the elastic term of \eqref{eq:eom} is
\begin{equation}
\begin{aligned}
f_{\mathrm{el},i}(x;\alpha) = \sum_{j\in\mathcal N(i)}\Big[
 &k_{\mathrm{lin}}(\alpha)(x_i-x_j) \\
 &+ k_{\mathrm{cub}}(\alpha)(x_i-x_j)^3\Big],
\end{aligned}
\label{eq:edge}
\end{equation}
where $\mathcal N(i)$ is the set of the four neighbors of node $i$, and $k_{\mathrm{lin}}$ and $k_{\mathrm{cub}}$ are the linear and cubic coefficients of the restoring force.
A node on the fixed rim has $x_j=0$.
Stiffness scales with $\alpha$ and the damping with $\sqrt{\alpha}$.
Collisions between the robot and the trampoline are handled by a two dimensional flex mesh of MuJoCo that covers the node network~\cite{mujoco_docs}.
Its vertices move with the node bodies and the contact forces are distributed to the nodes involved; the elasticity of the flex mesh itself is not used, and the restoring forces between nodes follow \eqref{eq:edge}.

\section{A Demonstration with Residual Reinforcement Learning}
\label{sec:method}

\begin{figure}[H]
\centering
\includegraphics[width=0.62\textwidth]{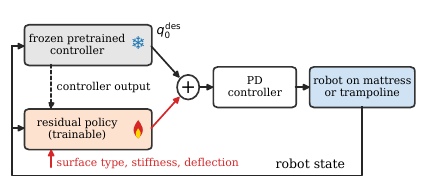}
\caption{Demonstration controller.
The pretrained network keeps its weights, a residual policy adds a bounded correction to its joint targets, and the sum goes to the original proportional derivative controller.
The residual policy reads the robot state and the output of the original controller.
The HoST residual also reads information about the surface.}
\label{fig:residual}
\end{figure}

Two pretrained controllers provide the original motion, and residual reinforcement learning trains an additional control term.
During training the original controller is frozen.
Its joint targets plus the residual output go to the original proportional derivative controller, which turns them into torque (Fig.~\ref{fig:residual}).

\textbf{Original controllers and residual parameterization.} Our first controller is the HoST \cite{huang2025learninghumanoidstandingupcontrol} standing policy.
It generates joint position increments from the robot state, and the released prone or supine policy is selected from the initial posture.
With $d$ controlled joints, current joint angles $q_t\in\mathbb R^d$ and HoST output $a^0_t\in\mathbb R^d$, the original joint target is $ q^{\mathrm{des}}_{0,t} = q_t + s_0\,a^0_t , $ where $s_0>0$ is the action scale of the original policy.
The second controller is the ProtoMotions \cite{tessler2024protomotions} motion tracker, whose deployment interface produces the joint target $q^{\mathrm{des}}_{0,t}$ from the robot state and a reference motion.
Let $o_t$ be the observation the residual policy receives at control step $t$.
During training the policy is a diagonal Gaussian whose output is bounded by a hyperbolic tangent,
\begin{equation}
\begin{aligned}
& a^{\delta}_t \sim \mathcal N\!\left(\mu_\theta(o_t),\operatorname{diag}(\sigma^2)\right), \\
& \delta_t = s\tanh(a^{\delta}_t),\\
& q^{\mathrm{des}}_t = q^{\mathrm{des}}_{0,t} + \delta_t ,
\end{aligned}
\label{eq:residual}
\end{equation}
where $a^{\delta}_t$ is the action sampled from the policy, $\delta_t$ is the residual correction added to the joint targets, and $q^{\mathrm{des}}_t$ is the joint target sent to the joint controller.
The transform acts componentwise, and $s>0$ is the bound on the correction of a joint target.

\textbf{Observation and task objective.} The observation of the residual policy contains the robot state and the output of the original controller.
For HoST it also contains information about the surface, and the second training stage adds the position of the robot in the task area (Table~\ref{tab:inputs}).
Let $z^{\mathrm{pelvis}}_t$ be the pelvis height in the world frame, $z^{\mathrm{sup}}_t$ the height of the top of the mattress cell or trampoline node directly below the lowest point of the collision geometry of the robot ($z^{\mathrm{sup}}_t=0$ on rigid ground), $R_{WB}(t)$ the rotation matrix from base coordinates to world coordinates and $e_z=(0,0,1)^\top$.
The uprightness of the base and the standing indicator are
\begin{equation}
\begin{aligned}
u_t &= e_z^{\top} R_{WB}(t)\, e_z ,\\
I_t &= \mathbf 1\{z^{\mathrm{pelvis}}_t - z^{\mathrm{sup}}_t \ge z_{\min},\; u_t \ge u_{\min}\},
\end{aligned}
\label{eq:standing}
\end{equation}
where $z_{\min}$ and $u_{\min}$ are the height and uprightness thresholds.
The pelvis is measured against the deformed support because on a soft mattress a robot standing upright sinks so far that its pelvis stays below $z_{\min}$ in the world frame.
A trial succeeds when the standing condition holds for $T_{\mathrm{hold}}$ control steps and the robot never leaves the task area.
The reward at each control step is
\begin{equation}
r_t = h_t + w_u\,u^{\mathrm{torso}}_t + I_t + w_J\,J_t - w_O\,O_t ,
\label{eq:reward}
\end{equation}
where $h_t\in[0,1]$ is the progress of the torso height from its initial value to the reference stand and $u^{\mathrm{torso}}_t\in[0,1]$ the uprightness of the torso, $J_t$ is one only at the step at which the trial first completes $T_{\mathrm{hold}}$ standing steps and zero otherwise, and $O_t$ is one when the robot leaves the allowed area, which also ends the training trial.
We train the residual policy with proximal policy optimization (PPO)~\cite{schulman2017proximalpolicyoptimizationalgorithms}.

\section{Evaluation Setup}
\label{sec:eval}

Table~\ref{tab:params} lists the parameters of the two surfaces, Table~\ref{tab:inputs} the inputs of the residual policy, and Table~\ref{tab:training} the controller, task, and training parameters.
For the mattress, the reference bed at $\alpha=1$ is a soft foam.
The cell stiffness $k_1=105.30$ N/m and the bending stiffness $b_1=50.72$ N/m are fitted to the force against indentation of two rigid flat discs of 356 mm and 100 mm diameter pressed into a finely discretized continuum foam model in Warp~\cite{macklin2022warp} with 6.25 mm elements, with a pooled RMS error of 1.7\%, and the cell damping $c_1=0.876$ N\,s/m gives a damping ratio of 0.25.
For the trampoline, the reference at $\alpha=1$ is derived from force versus displacement data digitized from Park et al.~\cite{park2023trampoline}.

HoST initial states follow the perturbation protocol of~\cite{huang2025learninghumanoidstandingupcontrol}, prone and supine, with 16 training and 48 test states per posture drawn independently, so each surface has 96 paired tests.
For ProtoMotions, the residual trains on 39 lying poses from the pose pool of HumanUP~\cite{he2025learninggettinguppoliciesrealworld}.
The 40 test poses are passive falls of the robot with its motors off on rigid ground from a perturbed standing pose.
Both controllers are evaluated on rigid ground, seven mattresses and two trampolines.

We implement the mattress and the trampoline a second time in Warp~\cite{macklin2022warp}, which shares no code with MuJoCo.
This second implementation is separate from the continuum foam model used for the calibration.
We compare the dynamic response after a steel ball is released at the surface center and its static deflection under different stiffness conditions.
When comparing the static deflection, we also use the static solution of the same spring law from Section~\ref{sec:backend} as a third reference, labeled analytic estimate in Fig.~\ref{fig:physics}.
Vlaovi\'c et al.~\cite{vlaovic2024mattress} pressed the loading pad of EN~1957~\cite{cen2012en1957} into a polyurethane mattress and recorded the load against indentation.
We fit the material scaling parameter to the shallow part of the curve and predict the load at the held-out indentation depths to further validate the physical accuracy of our simulation.

\section{Results}
\label{sec:exp}

\begin{figure}[!t]
\centering
\includegraphics[width=0.6\textwidth]{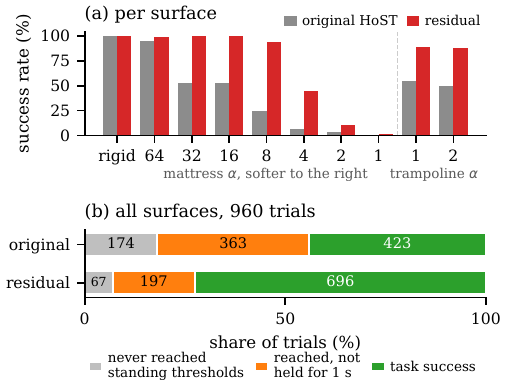}
\caption{HoST, 96 trials per surface.
\textbf{(a)} Success rate of the original controller and of the residual policy per surface.
\textbf{(b)} Outcomes over all the 960 trials in three classes.
The robot never reaches the standing thresholds, reaches them but either loses them before 1 s or leaves the task area, or succeeds.}
\label{fig:host}
\end{figure}

\begin{figure}[H]
\centering
\includegraphics[width=0.6\textwidth]{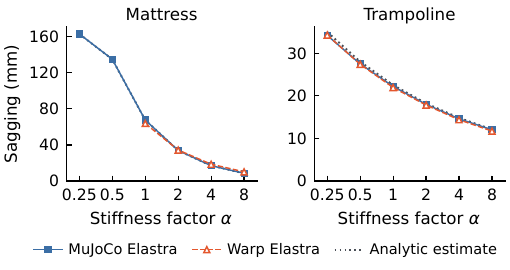}
\caption{Steel ball on the simulated mattress (left) and trampoline (right).
Static deflection versus stiffness in MuJoCo Elastra, in an independent Warp implementation, and from the analytic estimate.}
\label{fig:physics}
\end{figure}

\textbf{Systematic validation of the surface models.} As shown in Fig.~\ref{fig:physics}, the static deflections in MuJoCo Elastra and Warp Elastra agree within $0.3$~mm on the trampoline at all 6 stiffness levels and within $3$~mm on the 4 comparable mattress levels.
Both implementations agree with the analytic estimates.
Moreover, MuJoCo Elastra and Warp Elastra produce closely matching deflection curves during the initial sinking, rebound, and decaying oscillations.
The compression law also predicts the held-out loads of the real mattress~\cite{vlaovic2024mattress} with relative errors from $5.9\%$ to $10.8\%$.

\begin{figure}[!t]
\centering
\includegraphics[width=0.6\textwidth]{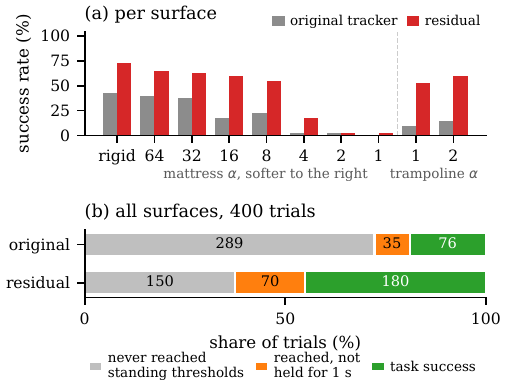}
\caption{ProtoMotions tracker on postures unseen in training, 40 trials per surface.
\textbf{(a)} Success rate of the original tracker and of the residual policy per surface.
\textbf{(b)} Outcomes over all the 400 trials in the three classes of Fig.~\ref{fig:host}.}
\label{fig:results}
\end{figure}

\textbf{HoST on mattresses and trampolines.} A residual policy recovers get-up on both the mattress and the trampoline (Fig.~\ref{fig:host}).
On rigid ground, both the original controller and the residual succeed in every trial.
On the three stiffer mattresses, the residual succeeds in 95 or 96 of 96 trials, and on the mattresses at $\alpha=8$ and $\alpha=4$ it raises success from $25\%$ to $94\%$ and from $6.3\%$ to $45\%$.
On the trampoline, the residual raises success from $55\%$ to $89\%$ at $\alpha=1$ and from $50\%$ to $88\%$ at $\alpha=2$.

\textbf{ProtoMotions on postures unseen in training.} The residual generalizes to fallen postures that were not used in training.
Fig.~\ref{fig:results} shows the ProtoMotions tracker on those postures, 40 trials on each mattress, trampoline and rigid ground, 400 trials in all.
Success rises from $19.0\%$ to $45.0\%$.
The largest gains are on the two trampolines, from $10\%$ and $15\%$ to $52.5\%$ and $60\%$.

\section{Conclusion and Future Work}
\label{sec:conclusion}

We built Elastra, simulated mattress and trampoline environments for robot learning, validated them in four ways, and demonstrated them with humanoid get-up by adapting frozen pretrained controllers with a small residual policy.
All robot experiments are simulated.
The surfaces are level and uniform, with the same stiffness and damping in every cell, whereas real supports vary from place to place, so nonuniform media are a natural extension.
The next steps are hardware experiments, nonuniform supports, other robots, and tasks such as a quadruped walking across a bed.

\section*{Acknowledgment}
Generative language models assisted in preparing this manuscript. GPT 5.6 Sol, Claude Opus 5, and Grok 4.6 were used to draft and revise prose in Sections~\ref{sec:intro} to \ref{sec:conclusion} and to help prepare tables and figure captions, and to write the simulation and training code of Sections~\ref{sec:backend} to \ref{sec:exp}. All claims, configurations and results were verified by the authors, who take full responsibility for the content of this paper.

\bibliographystyle{IEEEtran}
\bibliography{references}

\section*{Appendix: Parameters and Additional Figures}

\begin{table}[H]
\caption{Surface parameters used for training and evaluation with the robot.}
\label{tab:params}
\centering
\small
\setlength{\tabcolsep}{3pt}
\begin{tabular}{@{}p{3.0cm}p{6.2cm}p{5.8cm}@{}}
\toprule
 & Mattress & Trampoline \\
\midrule
Grid and size & $20\times19$ blocks of $0.10\times0.10\times0.05$ m covering $2.0\times1.9$ m & $25\times25$ nodes at 0.1 m spacing, circular region of radius 1.2 m \\
Vertical travel & 0.247 m, rigid floor 0.26 m below the reference top & 0.43 m \\
Elastic parameters & $k_1=105.30$ N/m, $b_1=50.72$ N/m, 306 bending terms, both $\propto\alpha$ & $k_{\mathrm{lin}}=3025.79\,\alpha$ N/m, $k_{\mathrm{cub}}=10^{-9}\alpha$ N/m$^3$ \\
Damping & $c_1=0.876$ N\,s/m, damping ratio 0.25, scales with $\sqrt\alpha$ & scales with $\sqrt\alpha$ \\
Inertial mass & 0.02915 kg per block, about 11.1 kg in all & node mass fixed \\
Contact geometry & rigid blocks moving with the cells & flex mesh moving with the nodes, collision radius 2 mm \\
\bottomrule
\end{tabular}
\end{table}

\begin{table}[H]
\caption{Inputs of the residual policy.
A factor is a fixed scaling applied before the input enters the network.}
\label{tab:inputs}
\centering
\small
\begin{tabular}{@{}p{6.0cm}rrp{6.0cm}@{}}
\toprule
Input & HoST & Tracker & Representation \\
\midrule
Joint positions & 23 & 29 & joint angles \\
Joint velocities & 23 & 29 & factor 0.1 \\
Base orientation & 4 & 4 & quaternion \\
Base angular velocity & 3 & 3 & base frame, factor 0.25 \\
Original controller output & 23 & 29 & HoST increment, tracker target \\
Torso height, torso uprightness, pelvis height & 3 & 3 & three scalars \\
Surface type & 3 & -- & one hot encoding \\
Stiffness factor & 1 & -- & $\log_2\alpha/6$, rigid ground 1 \\
Maximum surface deflection & 1 & -- & factor 10 \\
Robot position & 7 & -- & root $xy$, root velocity with factor 0.1, distances to the task area boundary; second training stage only \\
Progress of the trial & 1 & 1 & steps done over steps planned \\
\midrule
Total & 92 & 98 & \\
\bottomrule
\end{tabular}
\end{table}

\begin{table}[H]
\caption{Controller, task and training parameters.}
\label{tab:training}
\centering
\small
\setlength{\tabcolsep}{3pt}
\begin{tabular}{@{}p{3.0cm}p{12.6cm}@{}}
\toprule
HoST & 23 actuated joints, action scale $s_0=0.25$, observation history of six control steps, trial 5.02 s, correction bound 0.33 rad (1.32 action units) \\
ProtoMotions & 29 actuated joints, clips of 425, 361 and 497 control steps (8.50, 7.22, 9.94 s), heading aligned at the start, correction bound 0.85 rad \\
Time steps & simulation step 0.625 ms, control period 20 ms (32 substeps) \\
Standing condition & pelvis height above the support below the lowest robot point $\ge 0.7$ m, base uprightness $\ge 0.9$, held 50 control steps (1 s) \\
Reward weights & $w_u=0.5$, $w_J=20$, $w_O=10$ \\
Networks & policy and value, two hidden layers of 256 and 128, tanh, orthogonal init gain $\sqrt2$, policy output gain 0.01, initial log std $-1$ \\
PPO & 256 environments $\times$ 64 steps per update, 4 passes $\times$ 4 minibatches, 1000 updates for HoST in two stages of 500 and 500 updates for ProtoMotions (16,384,000 and 8,192,000 steps), $\gamma=0.995$, GAE 0.95, clip 0.2, lr $3\times10^{-4}$, value coef 0.5, entropy 0, grad norm 1.0, target KL 0.05 \\
Training surfaces & both controllers: rigid ground, mattress $\alpha\in\{8,16,32,64\}$ and trampoline $\alpha\in\{1,2\}$, mixed; second stage of HoST: mattress $\alpha\in\{1,2,4,8\}$ instead \\
Evaluation surfaces & both controllers on rigid ground, the mattress at $\alpha\in\{1,2,4,8,16,32,64\}$ and the trampoline at $\alpha=1$ and $2$ \\
\bottomrule
\end{tabular}
\end{table}

\end{document}